\documentclass[letterpaper]{article} 
\usepackage{aaai2026}  
\usepackage{times}  
\usepackage{helvet}  
\usepackage{courier}  
\usepackage[hyphens]{url}  
\usepackage{graphicx} 
\usepackage{natbib}  
\usepackage{caption} 
\usepackage{algorithm}
\usepackage{algorithmic}
\usepackage{amsmath}
\usepackage{amssymb} 

\usepackage{newfloat}
\usepackage{listings}
\DeclareCaptionStyle{ruled}{labelfont=normalfont,labelsep=colon,strut=off} 
\floatstyle{ruled}
\newfloat{listing}{tb}{lst}{}
\floatname{listing}{Listing}
\title{Finding the Translation Switch: Discovering and Exploiting the Task-Initiation Features in LLMs}
\author{
    Xinwei Wu\textsuperscript{\rm 1}\equalcontrib, 
    Heng Liu\textsuperscript{\rm 2}\equalcontrib, 
    Xiaohu Zhao\textsuperscript{\rm 2}, 
    Yuqi Ren\textsuperscript{\rm 1}, 
    Linlong Xu\textsuperscript{\rm 2}, \\
    Longyue Wang\textsuperscript{\rm 2}, 
    Deyi Xiong\textsuperscript{\rm 1}\thanks{Corresponding author.}, 
    Weihua Luo\textsuperscript{\rm 2}, 
    Kaifu Zhang\textsuperscript{\rm 2} 
}
\affiliations{
    \textsuperscript{\rm 1} TJUNLP, Tianjin University, China \\
    \textsuperscript{\rm 2} Alibaba International Digital Commerce, China\\
    \{wuxw2021, ryq20, dyxiong\}@tju.edu.cn\\
}

\usepackage{bibentry}

\begin{document}

\maketitle

\begin{abstract}
Large Language Models (LLMs) frequently exhibit strong translation abilities, even without task-specific fine-tuning. However, the internal mechanisms governing this innate capability remain largely opaque. To demystify this process, we leverage Sparse Autoencoders (SAEs) and introduce a novel framework for identifying task-specific features. 
Our method first recalls features that are frequently co-activated on translation inputs and then filters them for functional coherence using a PCA-based consistency metric. 
This framework successfully isolates a small set of ``translation initiation" features. 
Causal interventions demonstrate that amplifying these features steers the model towards correct translation, while ablating them induces hallucinations and off-task outputs, confirming they represent a core component of the model's innate translation competency.
Moving from analysis to application, we leverage this mechanistic insight to propose a new data selection strategy for efficient fine-tuning. 
Specifically, we prioritize training on ``mechanistically hard" samples—those that fail to naturally activate the translation initiation features. 
Experiments show this approach significantly improves data efficiency and suppresses hallucinations. 
Furthermore, we find these mechanisms are transferable to larger models of the same family. 
Our work not only decodes a core component of the translation mechanism in LLMs but also provides a blueprint for using internal model mechanism to create more robust and efficient models.
The codes are available at \url{https://github.com/flamewei123/AAAI26-translation-Initiation-Features}.
\end{abstract}

\section{Introduction}
Large Language Models (LLMs), particularly after instruction fine-tuning, often exhibit remarkable zero-shot translation capabilities, even without dedicated training on translation tasks \cite{hendy2023good,zhang2023prompting}. The origin of this emergent behavior remains a subject of debate. Early investigations posited that this ability stems from inadvertently included parallel corpora in the vast pre-training datasets \citep{li2024task}. However, subsequent work has shown that translation capabilities persist even when such data is meticulously removed \cite{briakou2023searching,fu2023reasonableness}, suggesting a more fundamental mechanism is at play. This data-centric approach to explanation has become increasingly untenable as pre-training corpora scale to trillions of tokens, making exhaustive analysis computationally prohibitive \cite{shen2023large,chang2024survey}. This leaves a fundamental question unanswered: what internal mechanism within instruction-tuned models is responsible for this innate translation ability?

\begin{figure}[t]
    \centering

    \includegraphics[width=0.85\columnwidth]{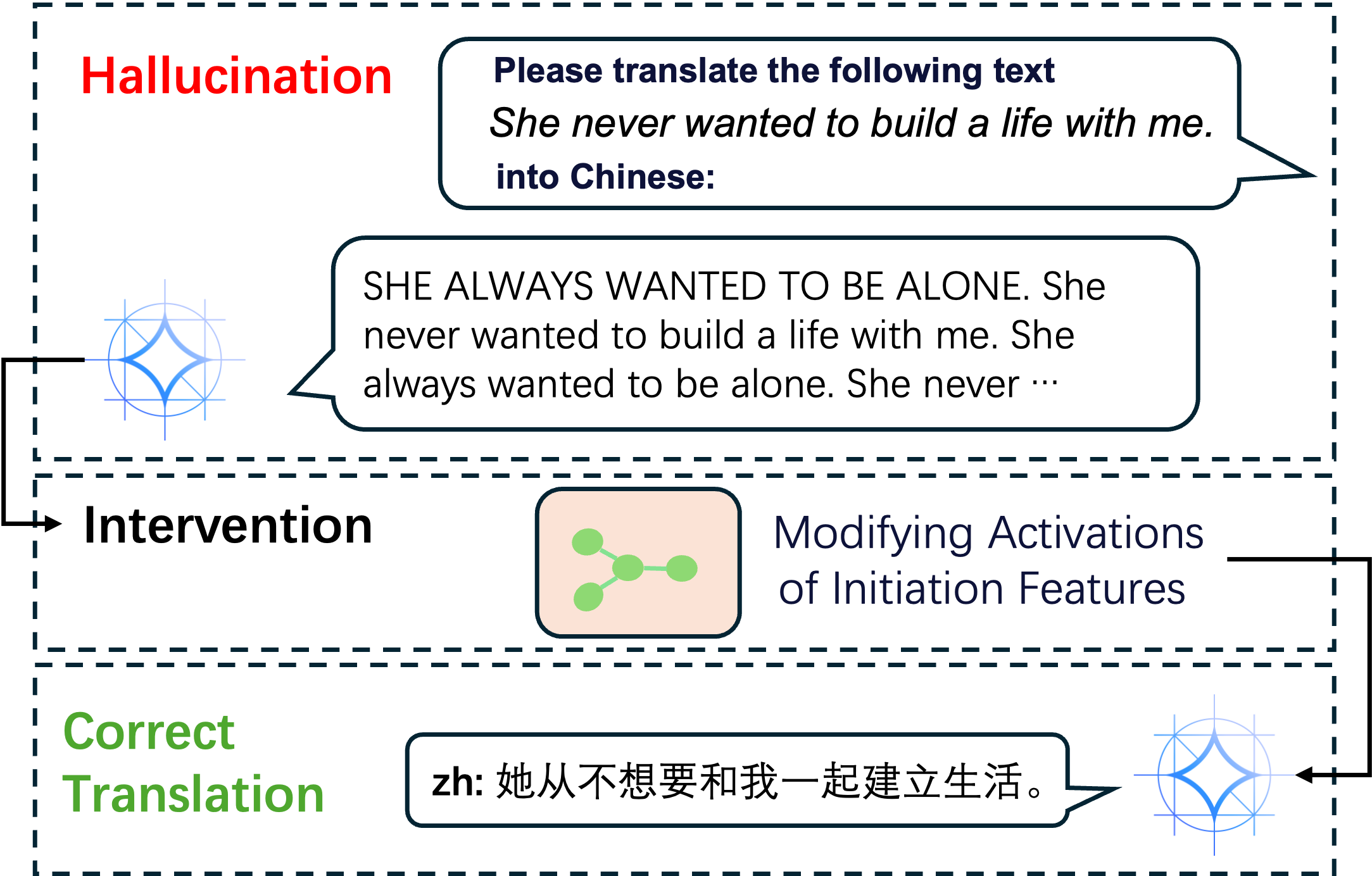} 
    \caption{An illustration of the effect of translation initiation features. Initially, the model hallucinates by failing to switch to the target language (top). By amplifying the identified ``translation initiation'' features (middle), we causally steer the model to produce the correct translation (bottom).}
    \label{fig:framework_illustration}
\end{figure}

In this work, we shift the focus from the pre-training corpus to the model's internal representations, employing Sparse Autoencoders (SAEs) to demystify the translation mechanism \cite{cunningham2023sparse,gao2024scaling,templeton2024scaling}. SAEs allow us to decompose a model's dense hidden states into a high-dimensional, yet sparsely activated, feature space, enabling more interpretable analysis of neural computations \citep{ cunningham2023sparse,lieberum2024gemma}. To this end, we propose a novel, three-stage framework for identifying task-specific features. First, we recall features that are frequently co-activated across critical prompt positions during translation tasks. Second, we derive a canonical, directionally stable feature influence vector for each candidate feature directly from the sparse representation of SAEs. Finally, we employ a PCA-based consistency metric to filter for a functionally coherent group of features whose vectors are highly aligned, representing a unified neural mechanism.

Applying this framework, we successfully isolate a small cluster of ``translation initiation'' features. Our analysis reveals two key properties of these features: (1) \textbf{Causal Generalization:} Causal interventions show that amplifying these features improves translation quality and reduces hallucinations \citep{wu2025challenging} across diverse language pairs. (2) \textbf{Mechanistic Function:} They steer the model to generate explicit translation-framing tokens, locking it into the correct task context. These findings provide compelling evidence for a dedicated, general-purpose circuit that controls the initiation of the translation task.

To demonstrate the practical utility of these mechanistic insights, we move beyond direct intervention, which incurs prohibitive computational overhead \cite{lishould, zhang2025explainable}. We propose a novel data selection strategy that uses the features' activation as an inner metric for ``instructive difficulty'', selectively fine-tuning on ``mechanistically hard'' samples where the model fails to naturally activate the initiation features. Our experiments demonstrate that this method significantly improves data efficiency and suppresses hallucinations. Furthermore, we show that these mechanistic insights are highly transferable to larger models \textit{within the same model family} (e.g., Gemma-2B to Gemma-9B), but do not generalize across different model architectures (e.g., Gemma to LLaMA).

Our contributions are threefold:
\begin{enumerate}
    \item We propose a novel, multi-stage framework for identifying functionally coherent, task-specific features in LLMs.
    \item We discover and analyze a ``translation initiation'' features, providing a mechanistic explanation for the model's innate translation abilities through causal and statistical evidence.
    \item We introduce a new mechanistic data selection method for fine-tuning, demonstrating that insights into internal model workings can lead to more efficient and robust model specialization.
\end{enumerate}

\section{Methodology}
\label{sec:methodology}

In this section, we detail our framework for identifying and filtering task-level features within LLMs. Specifically, our goal is to pinpoint the features responsible for initiating the translation task from the vast feature space decomposed by SAEs. Our methodology comprises three main stages: (1) \textbf{High-Frequency Feature Recall} to identify candidate features correlated with the task, (2) \textbf{Feature influence vector Characterization} to represent their directional influence, and (3) \textbf{Consistency-Based Filtering} to isolate functionally coherent and causally significant features. 

\subsection{Preliminaries: Sparse Feature Extraction via SAEs}
\label{subsec:preliminaries}

Our analysis focuses on the Gemma-2 model family \cite{team2024gemma}. We leverage the publicly available, pre-trained Sparse Autoencoders (SAEs) released by Google \cite{lieberum2024gemma}, which were trained on over 4 billion tokens. These SAEs are designed to decompose the model's dense hidden states into a more interpretable, sparse set of features.

Given a translation input to the LLM, we perform a forward pass to obtain the hidden states from a specific layer $l$. Let $\boldsymbol{h}^{(l)} \in \mathbb{R}^{d_{\text{model}}}$ be the hidden state vector at a particular token position from layer $l$, where $d_{\text{model}}$ is the dimension of the hidden state.

An SAE is trained to reconstruct this hidden state $\boldsymbol{h}^{(l)}$ from a sparse, high-dimensional feature activation vector $\boldsymbol{a}(\boldsymbol{h}^{(l)}) \in \mathbb{R}^{d_{\text{sae}}}$, where the feature dimension $d_{\text{sae}}$ is typically much larger than $d_{\text{model}}$. The SAE consists of an encoder and a decoder. The encoder, with weight matrix $\boldsymbol{W}_{\text{enc}} \in \mathbb{R}^{d_{\text{model}} \times d_{\text{sae}}}$, maps the hidden state to feature activations using the JumpReLU activation function, which is known to improve feature stability \cite{lieberum2024gemma}:
\begin{equation}
    \boldsymbol{a}(\boldsymbol{h}^{(l)}) = \text{JumpReLU}(\boldsymbol{W}_{\text{enc}}^T \boldsymbol{h}^{(l)} + \boldsymbol{b}_{\text{enc}})
    \label{eq:sae_encoder}
\end{equation}
The decoder, with weight matrix $\boldsymbol{W}_{\text{dec}} \in \mathbb{R}^{d_{\text{sae}} \times d_{\text{model}}}$, then attempts to reconstruct the original hidden state from these sparse activations:
\begin{equation}
    \hat{\boldsymbol{h}}^{(l)} = \boldsymbol{W}_{\text{dec}} \boldsymbol{a}(\boldsymbol{h}^{(l)}) + \boldsymbol{b}_{\text{dec}}
    \label{eq:sae_decoder}
\end{equation}
In this framework, we define a ``feature'' $f_{l,j}$ as a specific dimension $j$ of the SAE dictionary at layer $l$. Its activation for a given hidden state $\boldsymbol{h}^{(l)}$ is the $j$-th component of the sparse representation $\boldsymbol{a}(\boldsymbol{h}^{(l)})$. Each feature represents an abstract concept learned by the SAE. Our goal is to search through the tens of thousands of such features to find those causally linked to the translation task.

\subsection{Stage 1: Recalling High-Frequency Feature}
\label{subsec:recall}

A feature fundamental to translation should activate consistently when the model is prompted for the task \cite{geiger2024finding,deng2025sparse}. In our first stage, we filter for such features by identifying those that are frequently active across a corpus of $N$ translation samples. Recognizing that the task-related signal may appear at multiple locations, we monitor feature activations at three critical token positions: the final token of the source text (\texttt{src\_last}), the target language token (\texttt{tgt\_lang}), and the final query token (\texttt{input\_last}). 

For each sample, a feature $f_{l,j}$ is considered ``present'' if its activation is non-zero at the \textbf{union} of these three positions---that is, if it is active at \texttt{src\_last}, \texttt{tgt\_lang}, \textit{or} \texttt{input\_last}. We then recall all features that are present in at least 60\% of the samples ($\tau_{\text{freq}}=0.6$). This multi-position filtering strategy effectively narrows our search space to a manageable set of candidate features strongly correlated with the translation task.

\subsection{Stage 2: Characterizing Feature Influence Vectors}
\label{subsec:feature_vectors}

Stage 1 identifies features that are correlated with the translation task, but this does not guarantee their causal specificity or functional coherence \cite{deng2025sparse}. A truly task-specific feature should exert a consistent influence on the model's residual stream across different contexts.

To operationalize this, we define a \textbf{feature influence vector}, $\boldsymbol{v}_{l,j}$, which captures the causal effect of a feature $f_{l,j}$ within the context of a specific hidden state $\boldsymbol{h}^{(l)}$. This vector is computed by measuring the change in the SAE's reconstructed output when we force the activation of feature $f_{l,j}$ to a high value, $\alpha_{\text{act}}$, while keeping all other feature activations as determined by $\boldsymbol{h}^{(l)}$.

Formally, let $\hat{\boldsymbol{h}}_{\text{base}}$ be the standard reconstructed output from the SAE, and let $\hat{\boldsymbol{h}}_{\text{intervene}}$ be the output after the aforementioned intervention on feature $f_{l,j}$. The feature influence vector is defined as:
\begin{equation}
    \boldsymbol{v}_{l,j} \triangleq \hat{\boldsymbol{h}}_{\text{intervene}} - \hat{\boldsymbol{h}}_{\text{base}}
    \label{eq:feature_influence_vector}
\end{equation}
By this definition, each vector $\boldsymbol{v}_{l,j}$ is context-dependent. This allows us to proceed to our next stage: analyzing the directional consistency of these vectors across a diverse set of inputs to identify features with a stable, monosemantic function.

\subsection{Step 3: Consistency-Based Filtering}
\label{subsec:filtering}

The final stage validates whether our candidate features form a functionally coherent group. If a set of features truly represents a single underlying function, their corresponding feature influence vectors should be highly aligned, pointing in a near-identical direction in the residual stream. To quantify this collective alignment, we introduce a \textbf{PCA Consistency Score}.

Given a set of $n$ candidate feature influence vectors $\{\boldsymbol{v}_1, \boldsymbol{v}_2, \dots, \boldsymbol{v}_n\}$, our method first normalizes each vector to unit length, focusing purely on directional information. The PCA Consistency Score is then defined as the proportion of variance explained by the first principal component (PC1) of these unit vectors. This score measures how well a single direction can represent the entire set.

As implemented in our methodology, this score can be computed efficiently. Let $\boldsymbol{U} \in \mathbb{R}^{n \times d_{\text{model}}}$ be the matrix of the $n$ unit-normalized feature influence vectors. The score, $\rho$, is the largest eigenvalue of the covariance matrix $\boldsymbol{S} = \frac{1}{n}\boldsymbol{U}^T\boldsymbol{U}$.
\begin{equation}
    \rho = \lambda_{\text{max}}\left(\frac{1}{n}\boldsymbol{U}^T\boldsymbol{U}\right)
    \label{eq:pca_consistency}
\end{equation}
A score close to 1 indicates that the vectors are almost perfectly aligned, while a low score suggests functional divergence.

We apply this metric to the set of candidate features recalled in Stage 1. A set is deemed to represent a unified neural circuit and is selected only if its PCA Consistency Score exceeds a stringent threshold of $\tau_{\text{cons}} = 0.95$. This final step ensures we isolate a small, highly-curated group of features that are not only correlated with the task but are also functionally synonymous.

\section{Experiments}
\label{sec:experiments}

\subsection{Experimental Setup}
\label{sec:setup}

\subsubsection{Models and Interpretability Tools.}
Our interpretability analysis primarily targeted \textbf{Gemma-2-2B-IT} \cite{team2024gemma}. This model was selected due to the public availability of high-quality, pre-trained SAEs provided by Google \cite{lieberum2024gemma}. These SAEs were trained on the intermediate activations of the model's residual streams, which extended hidden states size from 2304 to 16384.

To assess the generalizability of our findings, we extended our causal analysis and fine-tuning experiments to two additional instruction-tuned models: \textbf{Gemma-2-9B-IT}, \textbf{LLaMA-3.1-1B-IT} and \textbf{LLaMA-3.2-8B-IT} \cite{dubey2024LLaMA}. These models, varying in architecture and scale, allow us to test the robustness of our conclusions across different model families. 

\subsubsection{Datasets and Partitioning.}
Our primary analysis utilized the \textit{WMT24++} dataset \cite{deutsch2025wmt24expandinglanguagecoverage}. To assess performance across diverse linguistic typologies, we select four translation directions: English-to-Chinese (en-zh), English-to-Arabic (en-ar), English-to-Russian (en-ru), and English-to-Japanese (en-ja). For each language pair, we used a subset of approximately 1,000 parallel sentences.

We used random samples of 98 sentence pairs as feature identification set. This set was used exclusively within our methodology (Section~\ref{sec:methodology}) to discover task-relevant features.
The remaining $\approx$900 sentence pairs were used for testing. This larger, held-out partition was used for all subsequent evaluations of translation performance and causal interventions.

Additionally, for the fine-tuning experiments presented in Section~\ref{sec:finetuning_application}, we constructed a larger data pool by randomly sampling 100,000 English-to-Chinese sentence pairs from the general \textit{WMT24} benchmark \cite{kocmi2024findings}. This dataset was used exclusively for training and evaluating our data selection strategies.

\subsubsection{Evaluation Metrics.}
We employ two core metrics to capture both the quality and faithfulness of the translations.
We use the \textbf{COMET} score \cite{rei2020comet}. COMET is a widely used, reference-based metric that has demonstrated a high correlation with human judgments.
Following the taxonomy proposed by \citet{huang2025survey}, we classify any translation that is unfaithful to the source text as a \textit{hallucination}. This includes instances of unexpected languages, empty output, and refusal to translate (e.g., repeating the source language). To quantify this, we adopt an LLM-as-a-Judge approach, closely following the evaluation setup proposed by \citet{gogoulou2025can}. The \textbf{Hallucination Rate} is defined as the percentage of translations judged as unfaithful by our judge model across the test set.

\subsection{R1: Results of Identifying Translation Features}
\label{sec:feature_results}

\subsubsection{Stage 1: High-Frequency Feature Recall.}
Following our methodology, we began by identifying an initial candidate set of features. We analyzed the activation patterns on our \textit{Feature Identification Set} within the MLP layers of Gemma-2-2B-IT and Gemma-2-9B-IT. Each layer's activations were decomposed into 16,384 sparse features by its pre-trained SAE. We flagged a feature as a candidate if it activates at the \textbf{union} of three critical prompt locations (\texttt{src\_last}, \texttt{tgt\_lang}, or \texttt{input\_last}) in over 60\% of the samples.

\begin{figure}[t!]
    \centering
    \includegraphics[width=0.9\columnwidth]{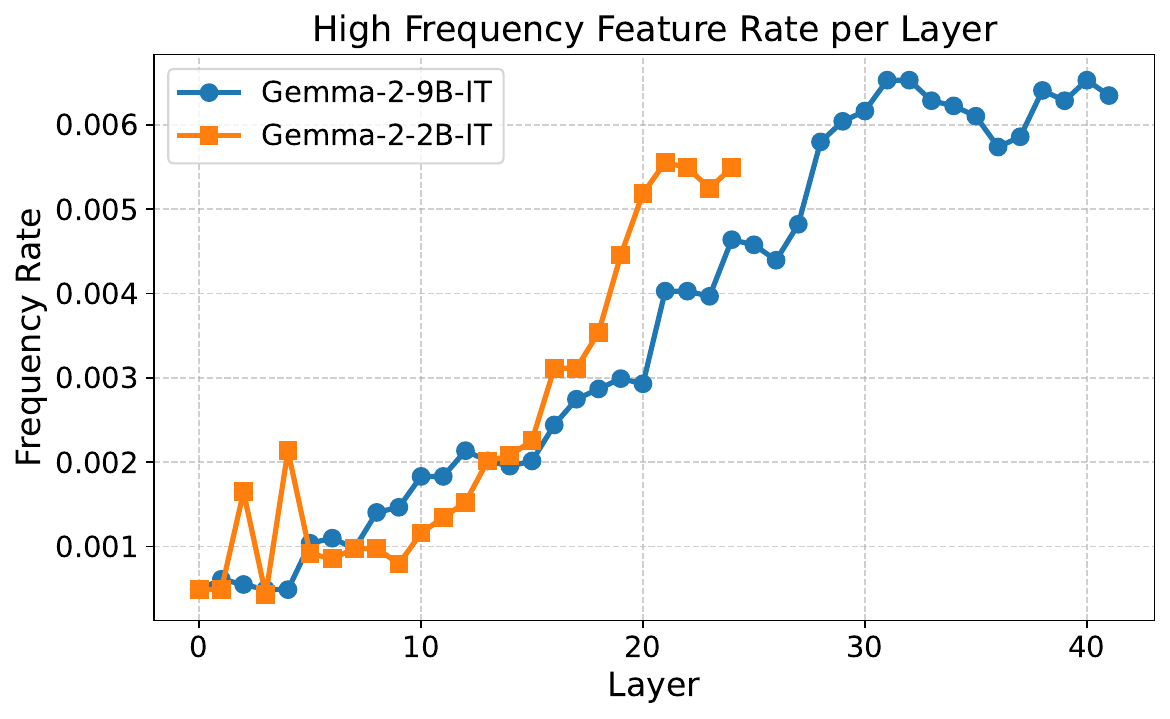} 
    \caption{Comparison of high-frequency feature rates per layer for Gemma-2-2B-IT and Gemma-2-9B-IT. }
    \label{fig:freq_recall_combined}
\end{figure}

The results of this recall stage are presented in Figure~\ref{fig:freq_recall_combined}. The figure plots the proportion of recalled features per layer and reveals a strong, shared pattern: the density of task-correlated features consistently increases with model depth. Despite the significant difference in scale (26 vs. 42 layers), the distributional shape is strikingly similar, pointing to a robust architectural principle for representing abstract tasks. This stage yields a total of 1,004 candidate features for Gemma-2-2B-IT and 2,485 for Gemma-2-9B-IT. 

\subsubsection{Stage 2 \& 3: Isolating a Coherent Feature set.}
The initial recall stage provides a broad set of candidates. To prune this set and isolate a functionally coherent set, we applied our PCA-based consistency filtering. We hypothesize that a true translation-initiation function is represented by a cluster of features whose canonical feature influence vectors are highly aligned. We tested this by grouping candidate features by layer and calculating a single \textbf{PCA Consistency Score} for each group, as defined in Equation~\ref{eq:pca_consistency}. A group is considered a coherent circuit only if its score exceeds $\tau_{\text{cons}} = 0.95$.

\begin{figure}[t!]
    \centering
    \includegraphics[width=0.95\columnwidth]{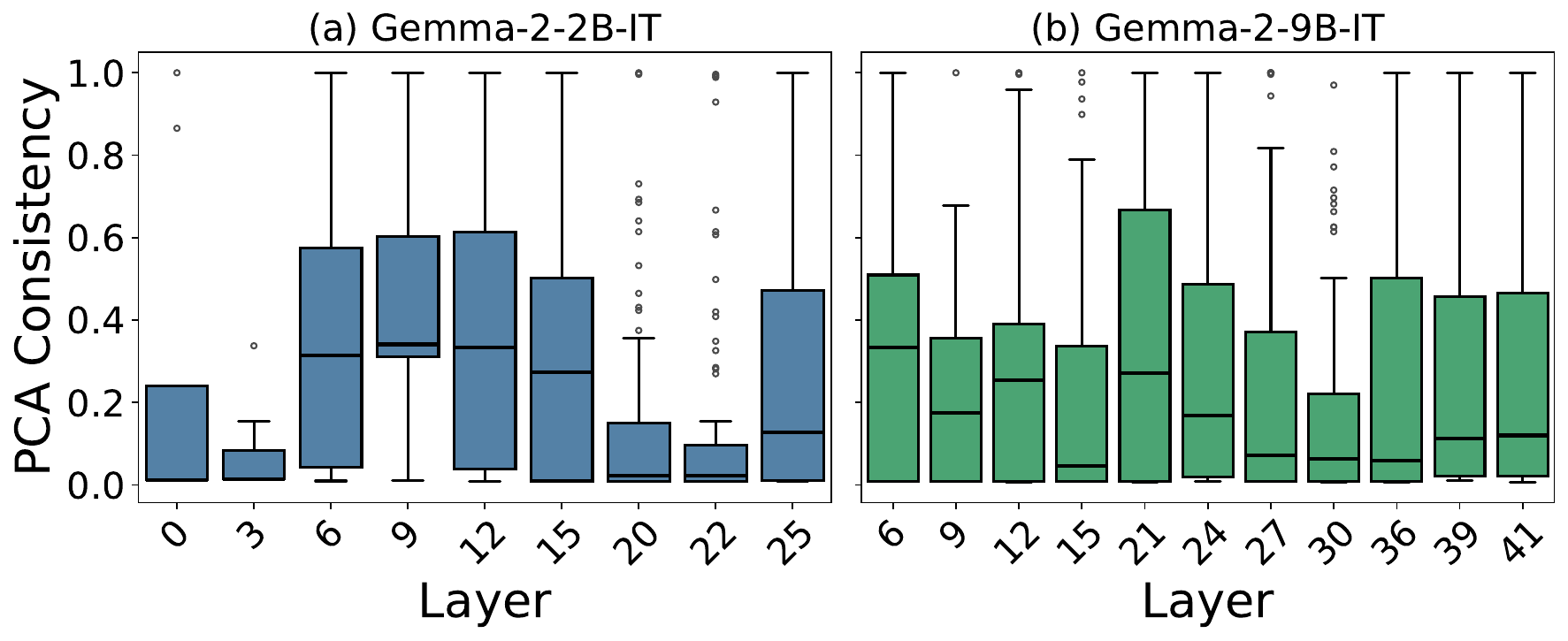} 
    \caption{Distribution of feature influence vector consistency scores for high-frequency features in (a) Gemma-2-2B-IT and (b) Gemma-2-9B-IT. }
    \label{fig:consistency}
\end{figure}

While our filtering operates at the group level, for finer-grained analysis, we can measure the alignment of each individual feature's vector with its group's principal component (PC1). Figure~\ref{fig:consistency} visualizes the distribution of these individual alignment scores. The boxplots reveal a critical insight: the majority of high-frequency features exhibit poor alignment, with median scores often below 0.4. This validates that co-activation alone is an insufficient criterion. However, a small number of outlier features achieve near-perfect alignment scores ($>0.95$). It is these features, residing within a handful of highly consistent layer-groups, that form our final, validated set of translation initiation features. For Gemma-2-2B-IT, this rigorous process distills the initial 1,004 candidates down to just \textbf{45} highly-consistent features. 

\subsubsection{Validating Causal Impact via Intervention.}
\label{sec:causal_impact}

To confirm that our consistency metric successfully isolates causally significant features, we conducted targeted intervention experiments on Gemma-2-2B-IT. On the en-zh test set, we intervened on features grouped by their individual alignment score: low ($<0.5$), medium ($[0.5, 0.95)$), and high ($>0.95$). We performed two types of \textbf{multiplicative interventions}:
\begin{itemize}
    \item \textbf{Ablation:} We multiplied a feature's activation by a coefficient of 0.
    \item \textbf{Amplification:} We multiplied a feature's activation by a coefficient of 2.0.
\end{itemize}
We then measured the resulting absolute change in COMET and Hallucination Rate to quantify each feature group's causal impact.

\begin{figure}[t!]
    \centering
    \includegraphics[width=0.98\columnwidth]{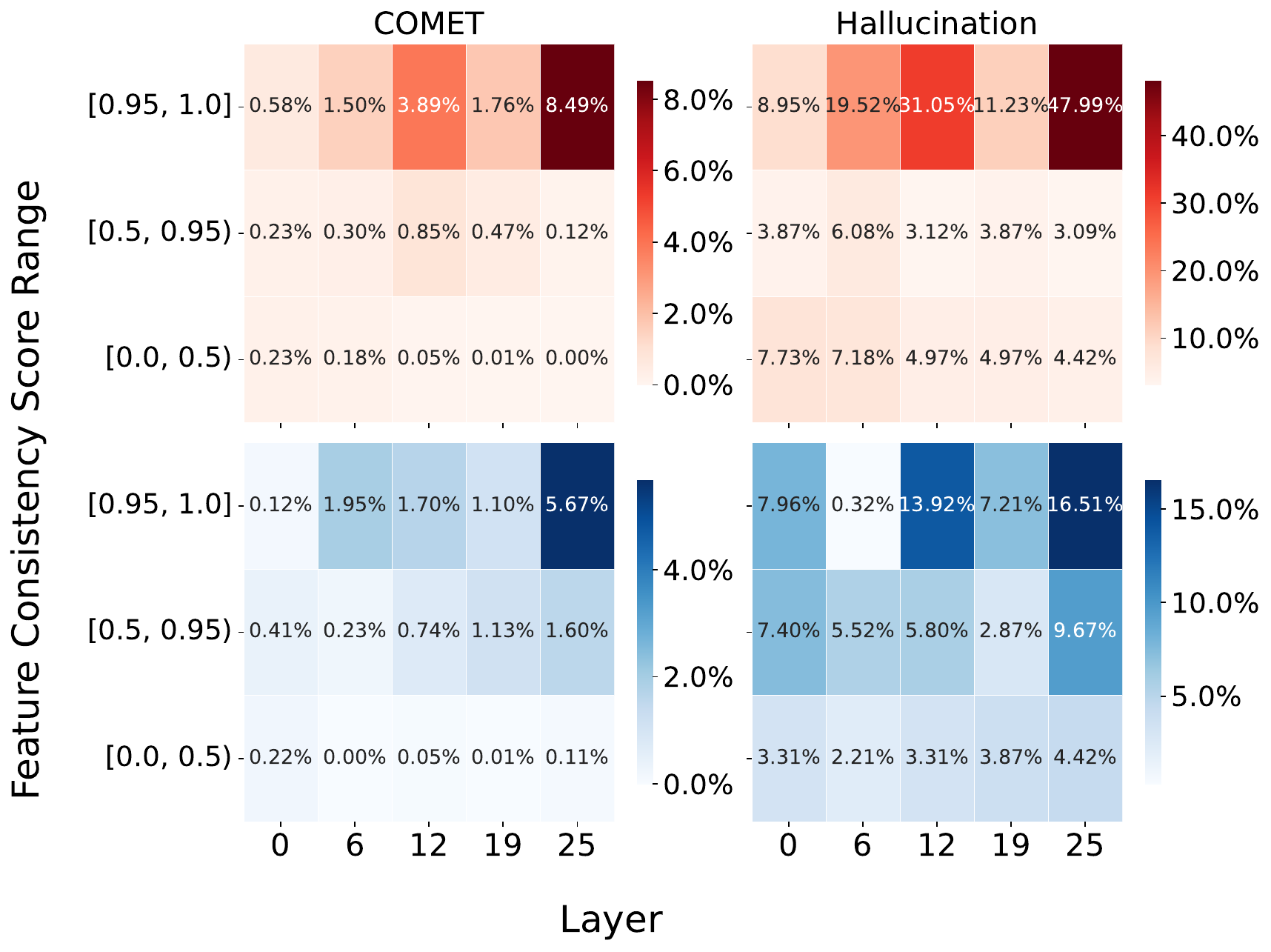}
    \caption{Absolute change in COMET and Hallucination Rate after feature intervention, grouped by layer and feature consistency score. Top row (red): ablation (coeff.=0). Bottom row (blue): amplification (coeff.=2.0).}
    \label{fig:causal_impact}
\end{figure}

The results, summarized in Figure~\ref{fig:causal_impact}, are decisive. Across all layers, a clear gradient emerges: \textbf{a feature's causal impact is directly proportional to its consistency score.} High-consistency features induce dramatic changes in model behavior, while low-consistency features have a negligible effect. For instance, ablating high-consistency features in layer 25 increases the hallucination rate by a staggering 47.99\% and degrades the COMET score by 8.49\%, confirming their necessity for faithful translation. This demonstrates that our framework successfully distills a large, noisy pool of candidates into a small set of features that act as the causal levers for the model's translation capabilities.

\subsection{R2: Analysis of Translation-task Features}

\begin{table*}[t!]
\centering
\begin{tabular}{l c c c c c c c c}
\hline
& \multicolumn{2}{c}{\textbf{en-zh}} & \multicolumn{2}{c}{\textbf{en-ja}} & \multicolumn{2}{c}{\textbf{en-ru}} & \multicolumn{2}{c}{\textbf{en-ar}} \\ \cline{2-3} \cline{4-5} \cline{6-7} \cline{8-9}
\textbf{Model } & \textbf{COMET $\uparrow$} & \textbf{Halluc. $\downarrow$} & \textbf{COMET $\uparrow$} & \textbf{Halluc. $\downarrow$} & \textbf{COMET $\uparrow$} & \textbf{Halluc. $\downarrow$} & \textbf{COMET $\uparrow$} & \textbf{Halluc. $\downarrow$} \\
\hline
Gemma-2-2B-IT & 73.62 & 19.15\% & 44.80 & 30.76\% & 54.36 & 29.46\% & 40.52 & 42.48\% \\
\quad + \texttt{l12-f2291} & \textbf{77.98} & 10.42\% & 47.62 & \textbf{17.89\%} & 55.59 & \textbf{16.37\%} & 42.02 & \textbf{29.47\%} \\
\quad + \texttt{l13-f3517} & 77.83 & \textbf{10.22\%} & \textbf{47.95} & 20.36\% & \textbf{57.20} & 19.26\% & \textbf{42.38} & 32.76\% \\
\hline
\end{tabular}
\caption{Cross-lingual generalization of translation-initiation features discovered on en-zh data. Activating feature \texttt{l12-f2291} or \texttt{l13-f3517} consistently improves COMET scores and drastically reduces hallucination rates across all tested language pairs, demonstrating their language-agnostic, task-initiating function.}
\label{tab:cross_lingual_generalization}
\end{table*}

\subsubsection{Cross-Lingual Generalization of Task Features.}
Our framework isolates a small set of high-impact features. We selected two prime candidates from Gemma-2-2B-IT, \texttt{l12-f2291} and \texttt{l13-f3517}\footnote{\texttt{l12-f2291} denotes the feature at position 2291 of layer 12.}, to test for functional generality. These features were discovered solely on English-to-Chinese (en-zh) data. To validate that they represent a language-agnostic function, we tested their causal effect on linguistically distant language pairs: En-Ja, En-Ru, and En-Ar. We performed a multiplicative intervention by amplifying each feature's activation during inference and evaluate the impact on COMET and hallucination rates.

The results, presented in Table~\ref{tab:cross_lingual_generalization}, demonstrate remarkable cross-lingual generalization. Amplifying either feature robustly improves translation performance across all four language pairs. Most notably, the intervention dramatically and universally slashes the hallucination rate, often by nearly half (e.g., from 19.15\% to 10.42\% in en-zh). The fact that features identified on a single language pair can universally enhance translation quality and faithfulness strongly suggests they do not encode language-specific knowledge. Instead, they encode a more abstract, language-agnostic instruction: to \textit{initiate translation}.

\subsubsection{Interpreting Features through Output Statistics}
\label{sec:interpreting_features}

\begin{figure}[t!]
    \centering
    \includegraphics[width=\columnwidth]{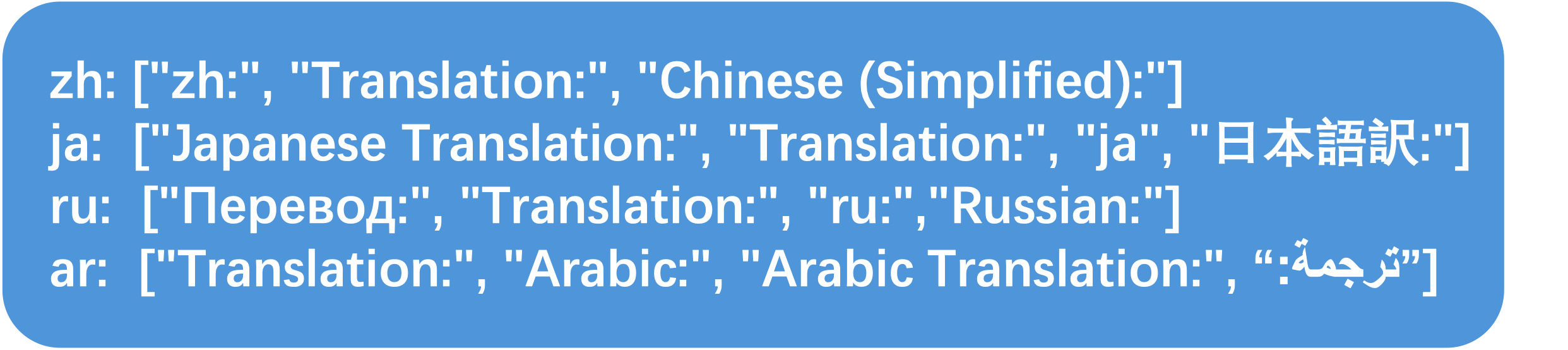}
    \caption{The curated lists of ``translation-framing tokens'' used for our analysis across the four target languages.}
    \label{fig:prompt_word_list}
\end{figure}

To elucidate the interpretable function of our identified features, we conducted a statistical analysis of the model's generated text. We first examined the baseline outputs and, through frequency analysis, observed that the model often uses a recurring set of tokens to preface its translations. We refer to these as ``translation-framing tokens,'' a sample of which is provided for each language in Figure~\ref{fig:prompt_word_list}.

\begin{figure}[t!]
    \centering
    \includegraphics[width=0.9\columnwidth]{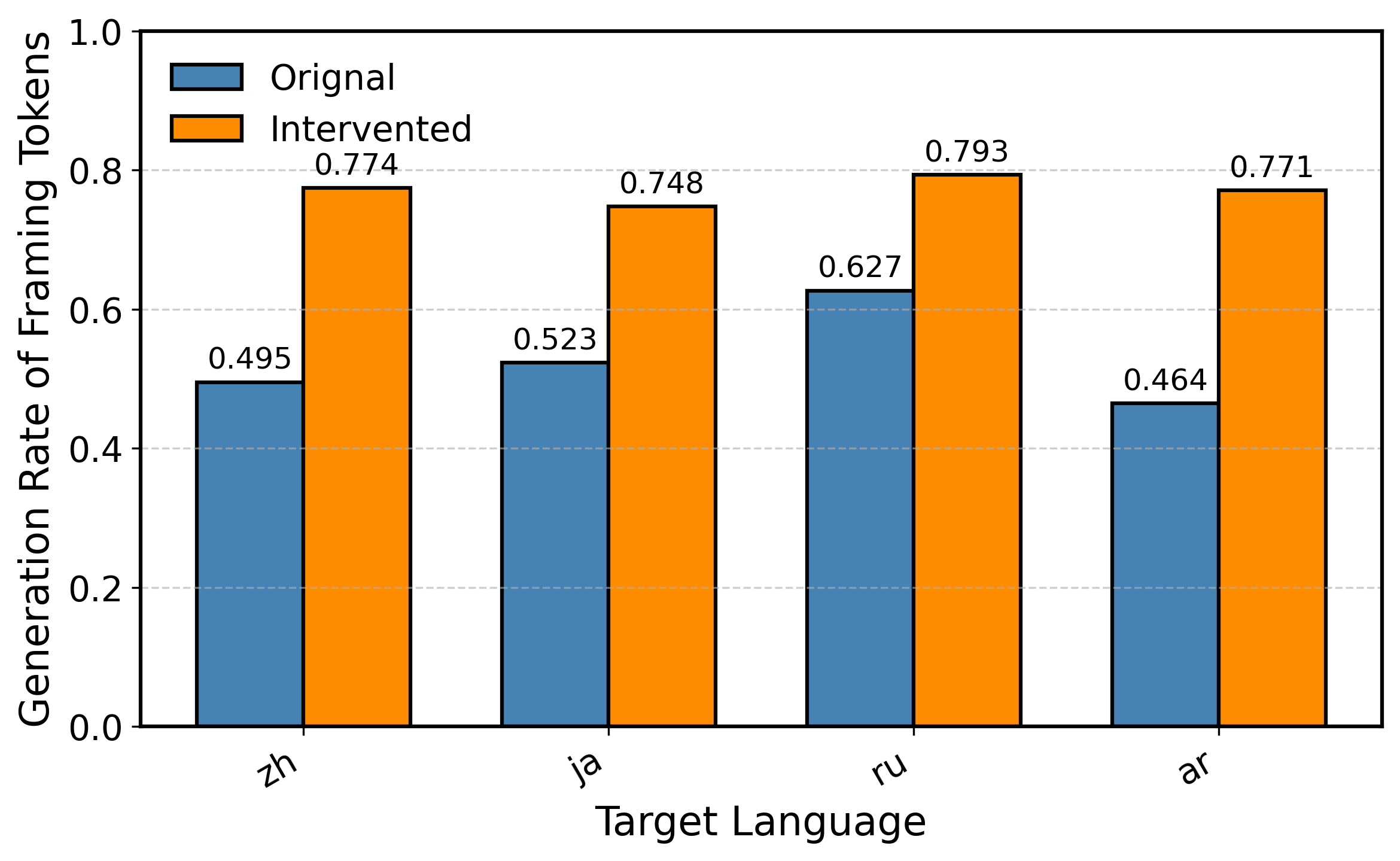}
    \caption{Rate of generating explicit ``translation-framing tokens'' with and without feature intervention. }
    \label{fig:prompt_word_rate}
\end{figure}

We then measured the generation frequency of these specific tokens, comparing the baseline model against outputs generated with our feature amplification intervention. The results, presented in Figure~\ref{fig:prompt_word_rate}, are unambiguous. Amplifying the feature's activation leads to a significant and consistent increase in the emission rate of these pre-identified framing tokens across all four languages. For instance, the rate for Arabic tokens surged from 46.4\% to 77.1\%.

This establishes a direct, observable link: the feature's activation is semantically correlated with an increased probability of generating these specific translation-framing tokens. This finding provides a clear, mechanistic explanation for the feature's function. By promoting the emission of tokens that explicitly initiate the task, the feature forces the model to commit to translation at the outset. This initial commitment preempts off-topic drift, providing a direct causal pathway to the observed reduction in hallucinations and the corresponding improvements in translation quality.

\subsection{Application: Mechanistic Data Selection for Efficient Fine-Tuning}
\label{sec:finetuning_application}

While direct causal intervention validates a feature's function, its computational overhead is impractical for deployment \citep{lishould, zhang2025explainable}. We therefore explore a more pragmatic application: using our identified feature as a signal for data selection \citep{xia2024less} to improve fine-tuning efficiency. Our rationale is that the initiation feature's activation reflects the model's innate ability to recognize the translation task. Samples that fail to activate this feature are ``mechanistically hard'' or confusing to the model. We hypothesize that selectively fine-tuning on these confusing samples make task initiation features robust, which means the inner metric can be an effective teaching signal rather than solely external metrics.

To test this, we conducted two sets of fine-tuning experiments. First, we validated the core principle of mechanistic selection by using a model's own feature signals for self-tuning. Second, we tested the transferability of these mechanistic insights by using features from a small model to select data for larger models. Additionally, we investigated the impact of different data selection ratios on the results.

\begin{table*}[t!]
\centering
\begin{tabular}{l l c c c}
\hline
\textbf{Model} & \textbf{Method} & \textbf{Data Count} & \textbf{COMET} $\uparrow$ & \textbf{Halluc. Rate} $\downarrow$ \\
\hline
\multicolumn{5}{l}{\textit{\textbf{Part 1: Self-Tuning Efficiency} (Using Model's Own Features)}} \\
\hline
\textbf{Gemma-2-2B-IT} & Original & - & 73.62 & 19.15\% \\
\cline{2-5}
& S0: Random & 20k & 82.49 & 3.62\% \\
& S1: High-Quality & 20k & 83.32 & 2.12\% \\
& S2: High-Loss & 20k & 82.14 & 4.32\% \\
& S3: Mechanistic (Ours) & 20k & \textbf{83.37} & \textbf{0.90\%} \\
\hline
\textbf{LLaMA-3.1-1B-IT} & Original & - & 57.61 & 32.24\% \\
\cline{2-5}
& S0: Random & 20k & 75.88 & 4.62\% \\
& S1: High-Quality & 20k & 76.35 & 4.21\% \\
& S2: High-Loss & 20k & 73.41 & 6.57\% \\
& S3: Mechanistic (Ours) & 20k & \textbf{77.92} & \textbf{2.39\%} \\
\hline
\multicolumn{5}{l}{\textit{\textbf{Part 2: Transfer-Tuning Scalability} (Using Features from Gemma-2-2B)}} \\
\hline
\textbf{Gemma-2-9B-IT} & Original & - & 79.50 & 12.50\% \\
\cline{2-5}
& S0: Random & 50k & 85.36 & 4.21\% \\
& S1: High-Quality & 50k & 84.17 & 4.97\% \\
& S2: High-Loss & 50k & 83.38 & 8.51\% \\
& S3: Mechanistic (Ours) & 50k & \textbf{86.48} & \textbf{0.60\%} \\
\hline
\textbf{LLaMA-3.2-8B-IT} & Original & - & 80.60 & 10.00\% \\
\cline{2-5}
& S0: Random & 50k & 86.20 & 0.20\% \\
& S1: High-Quality & 50k & \textbf{86.69} & \textbf{0.10\%} \\
& S2: High-Loss & 50k & 84.51 & 0.20\% \\
& S3: Mechanistic (Ours) & 50k & 86.34 & 0.30\% \\
\hline
\end{tabular}
\caption{Fine-tuning performance using different data selection strategies.  }
\label{tab:finetuning_results}
\end{table*}

\subsubsection{Part 1: Validating the Efficacy of Mechanistic Selection.}
In this experiment, we tested whether using a model's own internal signals can improve its fine-tuning efficiency. We fine-tuned Gemma-2-2B-IT and LLaMA-3.1-1B-IT on targeted subsets of a 100k en-zh data pool. For each model, we used its corresponding open-source SAE to identify its translation initiation feature. We then compared three data selection strategies:
\begin{itemize}
    \item (S0) \textbf{Random:} Selecting samples randomly.
    \item (S1) \textbf{High-Quality:} Selecting samples with the highest reference COMET score, which are generally high-quality for fine-tuning \citep{wang2024survey}.
    \item (S2) \textbf{High-Loss:} A hard-mining baseline selecting high-COMET samples that also incur the highest training loss \citep{cao2023instruction}.
    \item (S3) \textbf{Mechanistic (Ours):} Selecting high-COMET samples that exhibit the lowest activation of the model's own initiation features. We used the final set of high-consistency features produced by Step 3 of our framework. 
\end{itemize}

The results, detailed in the top half of Table~\ref{tab:finetuning_results}, confirm the effectiveness of our approach. For both Gemma-2-2B-IT and LLaMA-3.1-1B-IT, our mechanistic selection (S3) achieves a superior balance of performance, nearly matching the COMET scores of the best strategies while significantly reducing hallucination rates compared to other selection heuristics. This is particularly evident for Gemma-2-2B-IT, where our method achieves the performance of the full dataset with a fraction of the data, demonstrating high data efficiency. This confirms that a feature's activation level can be a reliable proxy for the ``instructive difficulty'' of a sample.

\subsubsection{Part 2: Testing the Transferability of Mechanistic Insights.}
A key question is whether mechanistic insights are portable across model scales and families. Here, we used the initiation feature identified in the small Gemma-2-2B-IT to select data for fine-tuning two larger models: Gemma-2-9B-IT (intra-family transfer) and LLaMA-3.2-8B-IT (cross-family transfer).

The results reveal a clear pattern of family-specific transferability.
\begin{itemize}
    \item \textbf{Intra-Family Transfer (Gemma $\rightarrow$ Gemma):} As shown for Gemma-2-9B-IT, the benefits of mechanistic selection are highly pronounced. Using data selected via the smaller model's feature allows the larger model to nearly match the performance of the full dataset. In contrast, the standard high-loss heuristic proves unreliable, leading to a substantial increase in hallucinations.
    \item \textbf{Cross-Family Transfer (Gemma $\rightarrow$ LLaMA):} The mechanistic insight does not transfer effectively. For LLaMA-3.2-8B-IT, the high-quality data baseline (S1) performs best, and our feature-based selection strategy offers no clear advantage.
\end{itemize}

These findings yield a crucial insight: while the underlying neural circuits for tasks like translation appear highly portable across different scales \textit{within the same model family}, they are not universally interchangeable across different architectural foundations. This demonstrates that our feature-based signal is a robust tool for efficient, family-specific model specialization.

\begin{figure}[t!]
\centering
\includegraphics[width=0.8\columnwidth]{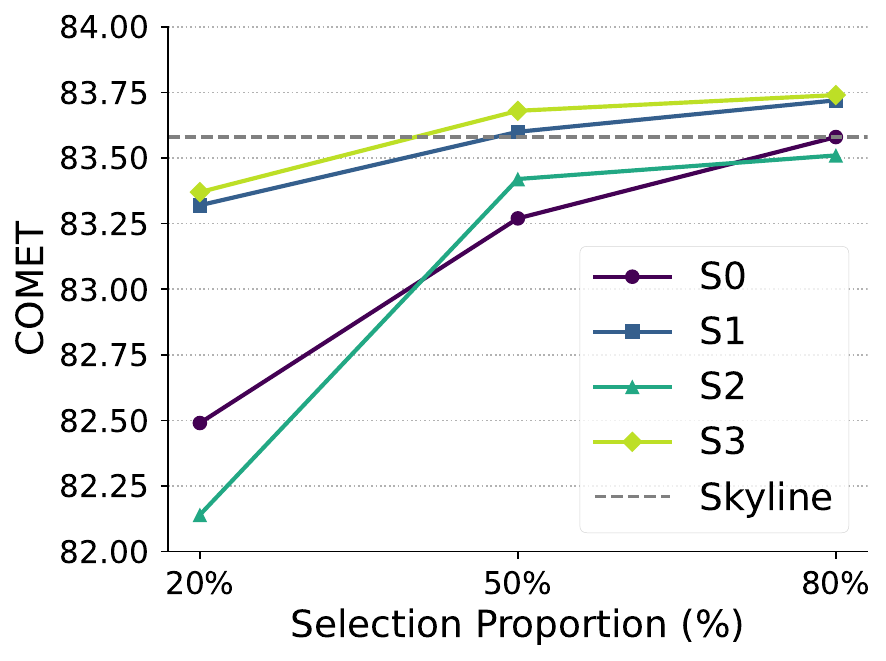} 
\caption{COMET score versus data selection proportion on the Gemma-2-2B-IT model.}
\label{fig:efficiency_curve}
\end{figure}

\subsubsection{Part 3: Impact of Data Selection Ratio.}
To further probe the data efficiency of our mechanistic selection, we analyzed how performance scales with the amount of selected data. We focused on the Gemma-2-2B-IT model and vary the fine-tuning data budget, selecting 20\%, 50\%, and 80\% of the 100k data pool using the four strategies (S0-S3). We benchmarked against a ``Skyline'' performance, which is the COMET score achieved by fine-tuning on the full 100k dataset (83.58).

The results, depicted in Figure~\ref{fig:efficiency_curve}, highlight the superior data efficiency of our mechanistic approach (S3). Across all data proportions, S3 consistently outperforms or matches the best-performing baselines, particularly excelling in the low-data 20\% regime. Most notably, with just 50\% of the data, our method's COMET score (83.68) \textbf{surpasses the full-dataset skyline} (83.58), demonstrating that it can curate a subset more effective than the entire dataset. This contrasts sharply with the high-loss (S2) and random (S0) baselines, which lag significantly. This finding strongly validates that our feature-based signal identifies a highly potent and compact training subset, maximizing performance while minimizing data requirements.

\section{Related Work}
\subsection{Large Language Models for Translation}

Large Language Models (LLMs) have surpassed traditional Neural Machine Translation (NMT) systems \cite{hendy2023good,li2023mmnmt}, leveraging document-level context \cite{sizov2024analysing,wang2023document} and world knowledge while offering superior controllability over style and linguistic variations via prompting \cite{zhang2023prompting,zhu2024towards}. Current specialization methods primarily involve Supervised Fine-Tuning (SFT) on parallel data \cite{zheng2024fine,wang2025marco} or Continual Pre-training to enhance multilingual foundations \cite{zhu2025survey,wang2025beyond}. 
Despite these advances, the innate translation ability of LLMs, which emerges even without explicit parallel training \cite{stap2024fine,ji2024zero}, remains a key area of investigation. A prevailing hypothesis has attributed this to ``incidental bilingualism'' within the pre-training corpus \cite{li2024task}, where non-parallel multilingual data inadvertently contains sentence-level translation pairs or code-switched text, providing implicit translation signals \cite{briakou2023searching}. 
However, such data-centric analyses are prohibitively expensive at scale. In contrast, our work leverages interpretability to probe the model's internal mechanisms directly. 

\subsection{Sparse Autoencoders for Interpretability}
\label{sec:sae_background}

Sparse Autoencoders (SAEs) are a key technique in mechanistic interpretability \cite{sharkey2025open} designed to address the superposition hypothesis \cite{elhage2022toy,gurnee2023finding}. 
This hypothesis posits that dense neural networks often encode multiple, distinct semantic concepts within the activations of a single neuron, making interpretation at both the neuron and embedding levels challenging.
SAEs mitigate this issue by decomposing these dense representations into sparse, more semantically meaningful features, thereby enhancing model transparency and interpretability \cite{cunningham2023sparse,gao2024scaling}.
Current research on SAEs generally falls into two categories. 
The first focuses on theoretical exploration and the automated identification of general computational mechanisms within models \cite{dunefsky2024transcoders}. 
For instance, SAEs have been used to analyze and reproduce the Indirect Object Identification (IOI) circuit \cite{makelov2024sparse} and to determine if a model is ``aware'' of specific knowledge entities \cite{ferrando2024know}. 
The second category focuses on identifying task-specific features, using their derived activation signals to precisely intervene on and steer model behavior \cite{templeton2024scaling,farrell2024applying,han2025towards}.
While prior SAE research has focused on simpler tasks, the complex mechanisms of translation remain underexplored. Our work addresses this gap by systematically applying SAEs to identify, validate, and leverage the key ``features'' within LLMs, extending mechanistic interpretability into a more practical and complex domain.

\section{Conclusion}

In this work, we have decoded a core component of the emergent translation capabilities in Large Language Models. We introduce a framework leveraging Sparse Autoencoders to isolate a ``translation initiation'' features and demonstrate through causal interventions that it governs the model's ability to perform translation faithfully. Moving beyond analysis, we translate this mechanistic insight into a practical data selection strategy that significantly improves fine-tuning efficiency and reduces hallucinations. Our findings also reveal that these circuits are portable across scales within a model family, but not across different architectures. This research provides a blueprint for how understanding the internal workings of LLMs can lead directly to more robust and efficient model specialization.

\section{Acknowledgments}

The present research was supported by the National Key Research and Development Program of China (Grant No. 2023YFE0116400). We would like to thank the anonymous reviewers for their insightful comments.

\bibliography{aaai2026}

\appendix

\section{Appendix}

\section{A. Experimental Settings}
\label{sec:appendix_settings}

This section provides supplementary details regarding the models, Sparse Autoencoders (SAEs), and key hyperparameters used in our experiments.

\subsection{Model and SAE Specifications}
Our experiments utilize several open-source models and their corresponding Sparse Autoencoders (SAEs), primarily sourced from the Hugging Face Hub. We detail the specifications for the two model families below.

For the Gemma model family, our primary analysis is conducted on \texttt{Gemma-2-2b-it}, which features a hidden size (\(d_{\text{model}}\)) of 2304 and 26 layers. For this model, we employ the \texttt{gemma-scope-2b-pt-res-canonical} SAE, which has a dictionary size (\(d_{\text{SAE}}\)) of 16384. For intra-family transfer experiments, we use \texttt{Gemma-2-9b-it} (\(d_{\text{model}}=3584\), 42 layers) as the target model.

Similarly, for the Llama model family, we use \texttt{Llama-3.2-1b-it} (\(d_{\text{model}}\) is 2048, 16 layers) for self-tuning validation, paired with the \texttt{chanind/sae-llama-3.2-1b-res} SAE (\(d_{\text{SAE}}\)) is 16384. The larger \texttt{Llama-3.1-8b-it} model (\(d_{\text{model}}\) is 4096, 32 layers) serves as the target for our cross-family transfer experiments.

\subsection{Hyperparameter Settings}

\subsubsection{Feature Identification Framework.}
The core thresholds for our feature identification pipeline were set as follows:
\begin{itemize}
    \item \textbf{Stage 1 (Recall):} A feature was flagged as a candidate if its activation frequency exceeded a threshold of \textbf{0.6} on our identification set.
    \item \textbf{Stage 3 (Filtering):} A layer-group of features was retained as a coherent circuit if its PCA Consistency Score surpassed a threshold of \(\tau_{\text{cons}} = \textbf{0.95}\).
\end{itemize}

\subsubsection{Causal Intervention.}
For our intervention experiments, we applied multiplicative scaling to feature activations during every step of the token generation process:
\begin{itemize}
    \item \textbf{Amplification:} The feature's activation value was multiplied by a coefficient of 2.0.
    \item \textbf{Ablation:} The feature's activation value was multiplied by a coefficient of 0.0.
\end{itemize}

\subsubsection{Instruction Fine-Tuning.}
The fine-tuning experiments for our data selection strategies were conducted using the hyperparameters detailed in Table~\ref{tab:appendix_ft_hyperparams}.

\begin{table}[!ht]
\centering
\caption{Hyperparameters for instruction fine-tuning experiments.}
\label{tab:appendix_ft_hyperparams}
\begin{tabular}{lc}
\hline
\textbf{Hyperparameter} & \textbf{Value} \\
\hline
Learning Rate & 1.0e-5 \\
Number of Epochs & 1.0 \\
FP16 Precision & True \\
Per-Device Batch Size & 64 \\
Gradient Accumulation Steps & 8 \\
\textbf{Effective Batch Size} & \textbf{512} \\
\hline
\end{tabular}
\end{table}

\begin{figure}[h!]
\centering
\fbox{
\begin{minipage}{0.95\columnwidth}
\small
\textbf{Irrelevant Translation:}
\texttt{
Your task is to assess the semantic relevance between the source text and its translation. \\
Source (\${source\_lang}): \${source\_text} \\
Target (\${target\_lang}): \${target\_text} \\
Determine if the target text is semantically unrelated to the source text. If the core meaning of the translation completely deviates from the source (i.e., it constitutes a 'hallucination' or is entirely off-topic), return '1'. If the translation maintains semantic correspondence with the source, even if imperfect, return '0'.
}
\end{minipage}
}
\end{figure}

\begin{figure}[h!]
\centering
\fbox{
\begin{minipage}{0.95\columnwidth}
\small
\textbf{Untranslated Content:}
\texttt{
Your task is to detect untranslated content in the target text. \\
Source (\${source\_lang}): \${source\_text} \\
Target (\${target\_lang}): \${target\_text} \\
Identify if any segment of the source text that requires translation has been left untranslated in the target text. Note that proper nouns, brand names, or specific terminologies might be intentionally retained, which should not be considered an error. An error occurs only when translatable content is incorrectly left in the source language. \\
If an untranslated error is detected, return '1'. Otherwise, return '0'.
}
\end{minipage}
}
\end{figure}

\begin{figure}[h!]
\centering
\fbox{
\begin{minipage}{0.95\columnwidth}
\small
\textbf{Repetition:}
\texttt{
Act as a language quality evaluator. Your task is to analyze the provided translation for redundancy issues. \\
Source (\${source\_lang}): \${source\_text} \\
Target (\${target\_lang}): \${target\_text} \\
Evaluate whether the target text contains unnecessary repetition of words or phrases that is not justified by the source text. If such erroneous repetition is present, return '1'. Otherwise, return '0'.
}
\end{minipage}
}
\end{figure}

\begin{figure}[h!]
\centering
\fbox{
\begin{minipage}{0.95\columnwidth}
\small
\textbf{Incorrect Language:}
\texttt{
Your task is to perform language identification on the provided text. \\
Input Text: \${target\_text} \\
Identify the language of this text and return its ISO 639-1 code (e.g., 'en' for English, 'pl' for Polish). If the language cannot be reliably determined, output 'unknown'.
}
\end{minipage}
}
\end{figure}

\section{B. Hallucination Definition and Detection Protocols}
\label{sec:appendix_hallucination}

\subsection{Definition of Hallucination}
In the context of Neural Machine Translation (NMT), hallucinations are traditionally defined as outputs that are unfaithful to the source text, such as fabricating content or introducing factual errors \citep{guerreiro2022looking, guerreiro2023hallucinations}. For this work, which focuses on general-purpose instruction-tuned LLMs rather than specialized NMT systems, we adopt a broader definition. Following the taxonomy of \citet{huang2025survey}, we categorize any failure to adhere to the translation instruction as a form of \textit{faithfulness hallucination}. This expanded definition includes not only semantic deviations but also instruction-following failures like producing output in the wrong language, refusing to translate, or generating irrelevant text.

\subsection{Hallucination Detection Prompts}
To systematically identify different types of hallucinations, we employed a suite of targeted prompts using an LLM-as-a-Judge model. Each prompt was designed to isolate a specific failure mode. The prompts are detailed below. Note that placeholders like \texttt{\${source\_lang}}, \texttt{\${source\_text}}, and \texttt{\${target\_text}} were populated programmatically for each sample.

\begin{description}
    \item[Irrelevant Translation] This prompt assesses whether the translation's core meaning deviates entirely from the source.
    \item[Untranslated Content] This prompt checks if translatable content was incorrectly left in the source language.
    \item[Repetition] This prompt identifies unnecessary repetition of words or phrases in the translation.
    \item[Incorrect Language] This prompt verifies if the output is in the correct target language.
\end{description}

\section{C. Case Studies}
\begin{figure*}[t!]
    \centering
    \includegraphics[width=0.8\textwidth]{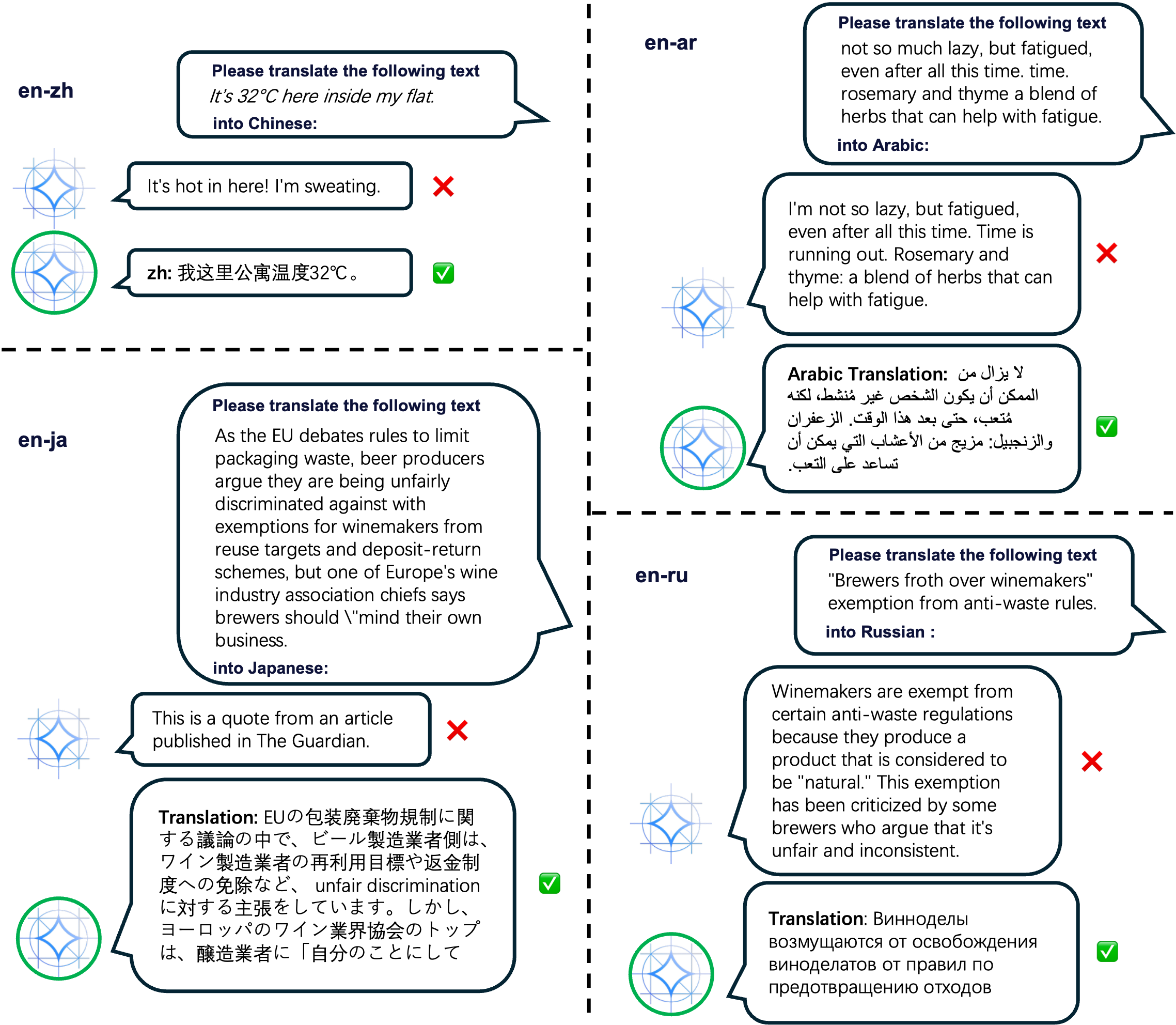} 
    \caption{
        Qualitative examples of causal intervention by amplifying a single translation initiation feature (\texttt{l12\_f2291}) in Gemma-2-2B-IT. For each of the four language pairs, the top dialog shows the input prompt. 
        The middle dialog, marked with a red Difference, shows the baseline model's output, which consistently exhibits instruction-following failures such as conversational replies (\textit{en-zh}), topic elaboration (\textit{en-ru} \& \textit{en-ja}), or source text continuation (\textit{en-ar}).
        The bottom dialog, marked with a green Checkmark, shows the output after our intervention. Amplifying the feature reliably steers the model to perform the correct translation task.
    }
    \label{fig:case_studies}
\end{figure*}

To provide a qualitative understanding of the "translation initiation" feature's causal role, we present several case studies in Figure~\ref{fig:case_studies}. The figure visually demonstrates how amplifying a single, identified feature (l12\_f2291 in Gemma-2-2B-IT) can correct various instruction-following failures and steer the model towards the correct translation task across multiple languages.

\end{document}